\documentclass[nohyperref]{article}

\usepackage{microtype}
\usepackage{graphicx}
\usepackage{subfigure}
\usepackage{booktabs} %
\usepackage{bm}
\usepackage{mathtools}
\usepackage[utf8]{inputenc} %
\usepackage[T1]{fontenc}    %
\usepackage{url}            %
\usepackage{booktabs}       %
\usepackage{nicefrac}       %
\usepackage{microtype}      %
\usepackage{graphicx}
\usepackage{multirow}

\usepackage{algorithm}
\usepackage{algorithmic}

\usepackage{hyperref}
\hypersetup{
    colorlinks,
    linkcolor={blue!50!black},
    citecolor={blue!50!black},
    urlcolor={blue!80!black}
}

\usepackage[preprint]{icml2023}

\usepackage[textsize=tiny]{todonotes}

\usepackage{subfigure}
\usepackage{amsmath,amsfonts}
\usepackage{amsthm,amssymb}
\usepackage{bbm} 
\usepackage[english]{babel}
\usepackage{xcolor}

\newcommand{\ie}{\emph{i.e.}}
\newcommand{\eg}{\emph{e.g.}}

\theoremstyle{plain}
\newtheorem{theorem}{Theorem}
\newtheorem{proposition}[theorem]{Proposition}

\newtheorem{observation}[theorem]{Observation}

\theoremstyle{definition}

\DeclareMathOperator*{\argmin}{arg\,min}

\newcommand*{\dif}{\mathop{}\!\mathrm{d}}
\newcommand{\KL}[2]{\mathcal{D}_{\mathrm{KL}}\left(#1\Vert #2\right)}

\newcommand{\N}{\mathcal{N}}
\newcommand{\D}{\mathfrak{D}}
\newcommand{\E}{\mathbb{E}}
\newcommand{\R}{\mathbb{R}}
\newcommand{\I}{\mathbf{I}}
\newcommand{\PP}{\mathbb{P}}

\newcommand{\x}{\mathbf{x}}

\newcommand{\z}{\mathbf{z}}
\newcommand{\s}{\mathbf{s}}
\newcommand{\f}{\mathbf{f}}
\newcommand{\w}{\mathbf{w}}
\newcommand{\dd}{{\boldsymbol{\delta}}}
\newcommand{\eps}{{\boldsymbol{\epsilon}}}
\newcommand{\vtheta}{{\boldsymbol{\theta}}}
\newcommand{\vphi}{{\boldsymbol{\phi}}}

\newcommand{\ltbc}{{\mathcal{L}_{\textrm{TBC}}}}

\def\Figref#1{Figure~\ref{#1}}

\def\Secref#1{Section~\ref{#1}}

\def\eqref#1{equation~\ref{#1}}
\def\Eqref#1{Equation~\ref{#1}}

\icmltitlerunning{Unifying Generative Models with GFlowNets and Beyond\hfill\thepage}

\begin{document}

\twocolumn[
\icmltitle{
Unifying Generative Models with GFlowNets and Beyond
}

\icmlsetsymbol{equal}{*}

\begin{icmlauthorlist}
\icmlauthor{Dinghuai Zhang}{mila}
\icmlauthor{Ricky T. Q. Chen}{fair}
\icmlauthor{Nikolay Malkin}{mila}
\icmlauthor{Yoshua Bengio}{mila,cifar}
\end{icmlauthorlist}

\icmlaffiliation{mila}{Mila -- Quebec AI Institute, Universit\'e de Montr\'eal}
\icmlaffiliation{fair}{Facebook AI Research (Meta AI)}
\icmlaffiliation{cifar}{CIFAR}

\icmlcorrespondingauthor{Dinghuai Zhang}{dinghuai.zhang@mila.quebec}
\icmlkeywords{Machine Learning, ICML}

\vskip 0.3in
]

\printAffiliationsAndNotice{}  %

\begin{abstract}

There are many frameworks for deep generative modeling, each often presented with their own specific training algorithms and inference methods. 
Here, we demonstrate the connections between existing deep generative models and the recently introduced GFlowNet framework~\citep{Bengio2021GFlowNetF}, a probabilistic inference machine which treats sampling as a decision-making process. This analysis sheds light on their overlapping traits and provides a unifying viewpoint through the lens of learning with Markovian trajectories. 
Our framework provides a means for unifying training and inference algorithms, and provides a route to shine a unifying light over many generative models. 
Beyond this, we provide a practical and experimentally verified recipe for improving generative modeling with insights from the GFlowNet perspective.
\end{abstract}

\section{Introduction}

Generative models are a class of machine learning algorithms that use probabilistic methods to capture and perform inference over complex distributions, usually from a given training dataset. 
They have a wide range of applications, including data generation,
anomaly detection, 
probabilistic inference and density estimation. In the past few decades, a variety of different generative models have been developed, each with its own set of assumptions and capabilities.

Early examples of generative models include probabilistic graphical models, such as Bayesian networks and Markov random fields~\citep{koller2009probabilistic,murphy2012machine}, and latent variable models, such as latent Dirichlet allocation~\citep{Blei2001LatentDA} and Helmholtz machines~\citep{Dayan1995TheHM}. These models have proven to be effective at capturing  dependencies within data.

The research on generative modeling has taken off during the past decade, thanks to the representational power of deep neural networks.
One well-known example is the generative adversarial network or GAN for short\citep{Goodfellow2014GenerativeAN}, which consists of a generator network that stochastically produces new samples and a deterministic discriminator network that tries to distinguish between real and generated samples. 
Another popular method is the variational autoencoder~\citep{Kingma2013AutoEncodingVB}, which learns a hierarchical latent variable model to express the target distribution with the help of a variational posterior.
Other types of generative models based on deep networks 
(deep generative models) 
include: normalizing flows~\citep{Dinh2014NICENI}, which transform a simple distribution into a target distribution using a series of invertible transformations; autoregressive (AR) models~\citep{Bengio1999ModelingHD,Oord2016PixelRN}, which model the distribution of a sequence of data by decomposing it into a product of conditional distributions; and energy-based models~\citep{Hinton2002TrainingPO,LeCun2006ATO}, which models the negative log probability of a distribution.
Recently, denoising diffusion models~\citep{Vincent2011ACB,SohlDickstein2015DeepUL,Ho2020DenoisingDP,Song2020ScoreBasedGM} have shown impressive results in generating high quality samples. Their modeling could be seen as a series of denoising steps which gradually transform white noise into noise-free data by stochastically inverting the process of transforming real data into white noise through a sequence of noise injection steps.
Each of these generative models has its own set of assumptions and limitations, which can make it challenging to choose the right model for a particular task~\citep{Hu2018OnUD}.

GFlowNets~\citep{Bengio2021FlowNB,Bengio2021GFlowNetF}, short for generative flow networks, is a class of samplers which stems from a reinforcement learning or RL for short~\citep{Sutton2005ReinforcementLA} formulation. GFlowNets treat the sampling process as a sequential decision-making process, and learn a stochastic (forward) policy to sample compositional objects with probability proportional to a given terminating reward function. It has been demonstrated that GFlowNets are able to sample from diverse modes rather than being stuck in single modes like is typical of Markov chain Monte Carlo (MCMC) or variational inference methods~\citep{Zhang2022GenerativeFN,Malkin2022GFlowNetsAV}, which is of great importance in drug discovery~\citep{Jain2022BiologicalSD,Zhang2021UnifyingLI}.

In this paper, we show how many existing generative models could be taken as special cases of GFlowNets, with their modeling components specified in different probabilistic ways. We thus propose to treat the GFlowNet as a probabilistic framework for unifying different kinds of generative models, and facilitate the analysis of their connections and possible extensions (\Secref{sec:unify_framework}). 
Further, after analysing the relationship between GFlowNet setup and generative modeling setup in \Secref{sec:gm_sampling}, we propose MLE-GFN, a generative modeling algorithm inspired by GFlowNet ideas
in \Secref{sec:mle_gfn_alg}. The proposed algorithm improves the performances of existing generative modeling baselines on both discrete and continuous image modeling tasks.

\section{Preliminaries and Notations}

\subsection{Generative Modeling}

Generative modeling aims to use probabilistic methods to model a distribution from a given dataset $\D=\{\x_i\}_i, \x_i\in \mathcal{X}$, where $\mathcal{X}$ is the space of data objects. When considering the training task, we want to learn a distribution $q(\x)$ to be close to the target distribution $p^{*}(\x)=\frac{1}{\vert\D\vert}\sum_i \delta_{\{\x_i - \x\}}$, where $\delta_{\{\cdot\}}$ is the Dirac distribution. However, the
real objective is to generalize well to
an underlying data generating process $p^*$
from which both training samples and test
samples may be drawn. One example to achieve this is to minimize the KL divergence, \ie, $\min_q\KL{p^{*}}{q}$, which corresponds to the maximum likelihood estimation (MLE), a popular method for training generative models. Other methods could also be taken as examples of divergence minimization, \eg, GAN's adversarial training could be viewed as minimizing the Jensen-Shannon divergence between the model and the target distribution. 

\subsection{GFlowNets}

From a probabilistic modeling viewpoint, a generative flow network (GFlowNet) \cite{Bengio2021FlowNB,Bengio2021GFlowNetF} is a probabilistic inference methodology that aims to sample $\x\in\mathcal{X}$ in proportion to a given reward function $R(\x)$, where $\mathcal{X}$ is the set of data. That is to say, the target distribution $p^*(\cdot)$ that we want so sample from satisfies $p^*(\x)\propto R(\x)$.
Recently, the community has experienced a progressive expansion of the concept of GFlowNets \cite{Malkin2022TrajectoryBI, Deleu2022BayesianSL, Jain2022BiologicalSD,Jain2022MultiObjectiveG,Pan2022GenerativeAF,Madan2022LearningGF,Liu2022GFlowOutDW}.
More precisely, a GFlowNet samples a Markovian trajectory $\tau=(\s_0, \s_1,\ldots, \s_{n})$ with length $n+1$, where $\s_i\in\mathcal{S}$ is the intermediate states for all $i$ in $[n]:=\{0, 1, \ldots, n\}$, whose space $\mathcal{S}$ is not necessarily the same as $\mathcal{X}$. If not specially specified, we use the notation $\x=\s_{n}$ for the final / terminating state of the trajectory.
This process has a natural connection to reinforcement learning~\citep{Sutton2005ReinforcementLA,Bengio2021FlowNB,Zhang2022LatentSM}.
The set of all trajectories $\tau$ form a directed acyclic graph (DAG) in the latent state space, whose nodes are states $\s \in \mathcal{S}$.
Each complete trajectory starts from the same (abstract) initial state $\s_0$ and ends in a terminating state $\s_n$.
The flow function $F(\tau) \in \mathbb{R}_{+}$ defined by \citet{Bengio2021GFlowNetF} can be understood by an analogy with the number of water particles flowing through trajectory $\tau$ in a network of pipes with $s_0$ as single source and all the terminating states as sinks.
Ideally, we want the amount of flow leading to $\x$ equals the given reward:
$\sum_{\tau=(\s_0, \cdots,\s_{n}), \s_n=\x}F(\tau) = R(\x)$.

Several training criteria have already been proposed for GFlowNets. We start from the {\bf flow matching} condition (\Eqref{eq:fm_condition}). Define $F(\s, \s')\triangleq \sum_{(\s\to \s')\in\tau}F(\tau)$  as the edge flow function and $F(\s)\triangleq \sum_{\s\in\tau}F(\tau)$ as the state flow function.
It is easy to see that the incoming flow into $\s'$ should match the outgoing flow from $\s'$:
\begin{align}
\label{eq:fm_condition}
\sum_{\s, \s\to\s'} F(\s, \s') = \sum_{\s'', \s'\to\s''} F(\s', \s''),
\end{align} 
as both side of the equation must equal $F(\s')$.
If one parametrizes the GFlowNet with the edge flow function $F_\theta(\cdot, \cdot)$, then the constraint in \Eqref{eq:fm_condition} would be used to define a corresponding training loss (\ie, $\mathcal{L}(\vtheta, \s')=\left(\log\sum_{\s} F_{\vtheta}(\s, \s') - \log\sum_{\s''} F_{\vtheta}(\s', \s'')\right)^2$), such that when the loss is zero everywhere, the constraint is satisfied, and 0 is also a global minimum of the loss.

The {\bf detailed balance} constraint of GFlowNets writes
\begin{align}
\label{eq:db_condition}
F(\s)P_{F}\left(\s' \mid \s\right) = F(\s')P_{B}\left(\s \mid \s'\right),
\end{align}
where $P_F(\s'|\s)$ and $P_B(\s|\s')$ are referred to as the forward and backward policy respectively and characterize the stochastic transitions between different states, going forward or backward along a trajectory. We can separately parametrize three models for $F_\vtheta(\cdot), P_F(\cdot|\cdot;\vtheta), P_B(\cdot|\cdot;\vtheta)$.
The detailed balance condition is closely related to the flow matching condition, in the sense that
$P_F(\s'|\s) = {F(\s, \s')}/{F(\s)}$ and $P_B(\s|\s') = F(\s, \s') / F(\s')$.
to determine a GFlowNet, it suffices to specify its forward policy \citep{Bengio2021GFlowNetF}.

Extending the detailed balance criterion to a constraint on the whole trajectory, the 
(general) {\bf trajectory balance} criterion~\citep{Malkin2022TrajectoryBI}
of GFlowNets aims to match GFlowNet's forward trajectory probability $P_{F}(\tau)$ and the backward trajectory probability $P_{B}(\tau)$, where
\begin{align}
P_{F}(\tau) &=\prod_{i=0}^{n-1} P_{F}\left(\s_{i+1}| \s_{i}\right) \\
P_{B}(\tau) &=\frac{R\left(\x\right)}{Z} \prod_{t=0}^{n-1} P_{B}\left(\s_{i}| \s_{i+1}\right),
\end{align}
where $\x=\s_n$. Notice that $\s_0$ is defined to be an abstract initial state (\ie, the first state of any trajectory) which has no concrete meaning for formalization reasons~\citep{Bengio2021GFlowNetF}.
Here $Z=\sum_\x R(\x)$ is the normalizing factor which generally needs to be learned (and can
then be included in the overall parameter vector $\theta$).
We also note that $P_{B}(\tau|\x)= \prod_{i=0}^{n-1} P_{B}\left(\s_{i} \mid \s_{i+1}\right)$ when $\x$ is the terminating state of the trajectory $\tau$.
Concretely, \citet{Malkin2022TrajectoryBI} proposes to use $(\log P_F(\tau)-\log P_B(\tau))^2$ as a training objective; nonetheless, in this work we focus on general divergences between the two distributions (\eg, KL divergence; we call such specification KL-trajectory balance).
We also use $P_T(\x)=\sum_{\tau=(\s_0, \cdots,\s_{n}), \s_n=\x}P(\tau)$ to denote the terminating distribution, namely the distribution on $\mathcal{X}$ that GFlowNet could generate.

\paragraph{Continuous GFlowNets.} \citet{continuous-gfn} developed a theory for GFlowNets with continuous or hybrid states, generalizing that for discrete states. In short, this theory allows one to use the same GFlowNet losses as in the discrete case, but replacing probability masses, such as $P_F(\s'|\s)$, by probability density functions $p_F(\s'|\s)$ as required, so long as all p.d.f.s are expressed in a compatible way, i.e., with respect to a common reference measure. In this paper, we are concerned with sampling in Euclidean spaces and will thus use $P_F$, $P_B$ interchangeably with their densities with respect to the Lebesgue measure on $\R^n$.

\section{GFlowNet as a Unifying Framework}
\label{sec:unify_framework}

\subsection{Hierarchical Variational Autoencoders}
\label{sec:hvae}

The evidence lower bound for bottom-up hierarchical VAEs (HVAEs) \cite{Ranganath2016HierarchicalVM} reads
\begin{align}
&\log p(\x) 
\ge \textrm{ELBO}_{p, q}(\x) \\
&\triangleq \E_{q(\z_{1:n-1}|\x)}\left[\log p(\x, \z_{1:n-1}) - \log q(\z_{1:n-1}|\x)\right] \\
&= \mathop{\E}_{q(\z_{n-1}|\z_{n})\cdots q(\z_1|\z_{2})}\left[\log p(\z_{1}) + \sum_{i=1}^{n-1}\log\frac{ p(\z_{i+1}|\z_{i})}{q(\z_{i}|\z_{i+1})}\right],
\end{align}
where we denote $\x:=\z_{n}$, $\z_{1:n-1}=(\z_1,\ldots,\z_{n-1})$, and $p, q$ respectively denote the hierarchical decoder and encoder of the HVAE. 
It is well known that this hierarchical ELBO can also be represented as
$
\KL{q(\z_{1:n-1}|\x)}{p(\z_{1:n-1}|\x)},
$
where $p(\z_{1:n-1}|\x)\propto p(\x|\z_{1:n-1})p(\z_{1:n-1})$.
As we show below, with a GFlowNet that samples $\z$ given $\x$ and where the reward is $p(\x, \z_{1:n-1})$, we also aim to match the forward trajectory policy which ends with data $\x$ with the corresponding backward trajectory policy, \ie, $P_F(\tau)\approx P_B(\tau)$, conditioning on the event $\{\x\in\tau\}$, \ie, $\x$ is the terminating state of $\tau$. Note that we have $\PP(\x|\tau)=\delta_{\{\x\in\tau\}}$, where $\PP(\cdot)$ denotes the probability of some event.

\begin{observation}
\label{obs:hvae_gfn}
The HVAE is a special kind of GFlowNet in the following sense:
each trajectory is of the form of $\tau = (\z_0, \z_{1},\cdots, \z_{n}=\x)$. 
The VAE decoder, which samples 
$\z_0\to\cdots\to\z_{n-1}\to\z_{n}$,
corresponds to the GFlowNet forward policy;  
the encoder samples 
$\x\to\z_{n-1}\to\cdots\to\z_0$,
and corresponds to the GFlowNet backward policy.
\end{observation}

With $\x=\z_{n}$, we could then write
\begin{align}
\label{eq:hvae_decoder}
P_F(\tau) = p(\z_0)\prod_{i=1}^{n-1} p(\z_{i+1}|\z_i), \\
\label{eq:hvae_encoder}
P_B(\tau) = p^*(\x) \prod_{i=1}^{n-1} q(\z_i|\z_{i+1}),
\end{align}
where $p^*(\x)=\frac{R(\x)}{Z}$ is the true target density, $p(\cdot|\cdot)$ is the forward policy, and $q(\cdot|\cdot)$ is the backward policy.
We can see that \textit{the $i$-th state in a GFlowNet trajectory (\ie, $\s_i$) corresponds to $\z_i$}.

The following proposition reveals an equivalence between the two perspectives in Observation~\ref{obs:hvae_gfn}.
\begin{proposition}
\label{prop:hvae_kl_elbo}
    Training hierarchical latent variable models with the KL-trajectory balance
    $\KL{ P_B(\tau)}{P_F(\tau)}$ objective
    is equivalent to training HVAEs by maximizing its ELBO, in the sense of having the same global optimum.
\end{proposition}
In this work we relegate all proofs to \Secref{sec:proofs}. We refer to \citet{Snderby2016LadderVA,Child2021VeryDV, Shu2022BitPI} for practice of HVAEs on large scale image modeling.

\subsection{Diffusion Models}
\label{sec:diffusion}

\subsubsection{Denoising diffusion probabilistic models}

The success of deep learning relies on careful design of inductive biases in the learning algorithm~\citep{Goyal2020InductiveBF}. With the same amount of computational resources and data, the better assumptions we use to constrain the model, the more powerful and better generalizing the algorithm will be. One way to bake inductive biases into the aforementioned hierarchical VAE model is by forcing the following Gaussian assumptions:
\begin{align}
\label{eq:ddpm_pf}
P_F(\s_{i+1}|\s_i) &= \N(\s_{i+1}; \boldsymbol{\mu}_i(\s_i), \beta_i\I), \\
P_B(\s_i|\s_{i+1}) &= \N(\s_i; \sqrt{1-\beta_i}\s_{i+1}, \beta_i \I),
\label{eq:ddpm_pb}
\end{align}
where $i=0,\ldots, n-1$ are integer indices, $\N$ denotes the Gaussian distribution, $\{\mu_i(\cdot)\}_i$ are functions to be learned, and $\{\beta_i\}_i$ are constant positive real numbers.
In this way, the latent variables at every stage of the sequential generative process share the same number of dimensions as data $\x$. 

\begin{observation}
\label{obs:ddpm}
    The denoising diffusion probabilistic model or DDPM for short~\citep{Ho2020DenoisingDP} is a special kind of GFlowNet with the forward policy\footnote{There is another specification of the DDPM generative decoder variance, which we ignore as it does not affect our discussion.} specified as in \Eqref{eq:ddpm_pf} and the backward policy specified as in ~\Eqref{eq:ddpm_pb}.
\end{observation}

Remark that our notation of the time index is in the reverse ordering of the one used in the DDPM exposition, which results in a slightly different definition of $\beta$. With the special design in \Eqref{eq:ddpm_pf} and ~\ref{eq:ddpm_pb}, a DDPM enjoys an efficient training procedure (each $\boldsymbol{\mu}_i$ can be trained locally by only trying to invert the noise added to $s_{i+1}$ to obtain $s_i$), which helps it scale well and made it become a state-of-the-art method for high-dimensional image generative modeling~\citep{Dhariwal2021DiffusionMB, Kingma2021VariationalDM}. We analyze the relationship between its training objective and a GFlowNet formulation in the following proposition.

\begin{proposition}[informal]
    \label{prop:ddpm_elbo}
    Training a GFlowNet defined as in \Eqref{eq:ddpm_pf} and ~\ref{eq:ddpm_pb} with KL-trajectory balance is equivalent to training a DDPM with its regression-based denoising objective.
\end{proposition}

\paragraph{Discrete space diffusion} 
The above modeling method could also generalize to structured discrete data. For categorical (\ie, discrete) data, $\s_i$ denotes the one-hot representation of $\s_i$, and we define a hierarchical model as in \Secref{sec:hvae} in the manner of the following claim, which parallels Observation~\ref{obs:ddpm}:

\begin{observation}
The discrete denoised diffusion model for categorical data proposed by \citet{Hoogeboom2021ArgmaxFA,Austin2021StructuredDD} is a special case of GFlowNet with the forward policy specified in \Eqref{eq:d3pm_pf} and backward policy specified in \Eqref{eq:d3pm_pb}.
\begin{align}
    P_F(\s_{i+1}|\s_i) &= \mathcal{C}(\s_{i+1};\mathbf{p}=
    \boldsymbol{\mu}_i(\s_i)), \label{eq:d3pm_pf}\\ 
    P_B(\s_i|\s_{i+1}) &= \mathcal{C}(\s_{i}; \mathbf{p}=\s_{i+1}\mathbf{Q}_{i}),  \label{eq:d3pm_pb}
\end{align}
where $\mathcal{C}$ denotes a categorical distribution, $\{\mathbf{Q}_i\}_i$ are doubly-stochastic constant Markov transition matrices, and $\{\boldsymbol{\mu}_i(\cdot)\}_i$ are parametric functions serving as the categorical parameter for the GFlowNet forward policy.
\end{observation}

\subsubsection{Denoised diffusion through SDEs}

The behavior of a deep latent variable model in its infinite depth regime is studied by~\citet{Tzen2019NeuralSD};
in the language of GFlowNets, the forward and backward policy take the following form: 
\begin{align}
P_F(\s_{i+1}|\s_i) &= \N\left(\s_{i+1}; \s_i + h\f_i(\s_i), hg^2_i)\right),\\
P_B(\s_i|\s_{i+1}) &= \N(\s_i; \s_{i+1}+h\large(-\f_{i+1}(\s_{i+1}) \\
&\quad\quad + g^2_{i+1}\nabla\log 
F
(\s_{i+1})\large) ,hg^2_{i+1}),
\end{align}
where we assume all $\s_i$ have the same number of dimensions as $\x$, time step $h=1/L$, $\{g_i\}_i$ are scalar parameters, and $\{\f_i(\cdot): \mathcal{S}\to\mathcal{S}\}_i$ are parametric mappings.
The hierarchical model is then equivalent to a stochastic process in its diffusion limit (
$h\to 0$).
\citet{Huang2021AVP, Kingma2021VariationalDM, Song2021MaximumLT} also study connections between hierarchical variational inference on deep latent variable models and diffusion processes.

Consider a stochastic differential equation (SDE) \cite{ksendal1985StochasticDE} and its reverse time SDE \cite{Anderson1982ReversetimeDE}
\begin{align}
\label{eq:sde}
\dif \x &= \f(\x)\dif t + g(t)\dif \w_t,   \\
\label{eq:reverse_sde}
\dif \bar\x &= \left[\f(\bar\x)-g^2(t) \nabla_{\bar\x}\log p_t(\bar\x)\right]\dif \bar{t} + g(t)\dif \bar\w_t,   
\end{align}
where $\f(\cdot)$ and $g(\cdot)$ are given and $\x, \w\in\mathbb{R}^D$, $\dif\w_t$ is a Wiener process,  and $\bar\x, \bar{t}, \bar\w_t$ denote the reverse time version of $\x, t, \w_t$.
We define $\PP_h(\x_{t+h}|\x_t)$ to be the transition kernel induced by the SDE in \Eqref{eq:sde}, namely
$\x_{t+h} = \x_t + \int_{t}^{t+h}f(\x_\tau) \dif\tau + \int_{t}^{t+h}g(\tau)\dif\w_\tau$, where $h$ denotes an infinitesimal time step.
This modeling, adopted and popularized by \citet[ScoreSDE]{Song2020ScoreBasedGM}, could be connected to GFlowNets as follows.

\begin{observation}
ScoreSDE is a special case of GFlowNets, in the sense that GFlowNet states take the time-augmented form $\s_t=(\x_t, t)$ for some $t\in [0,1]$,
the SDE in \Eqref{eq:sde} models the forward policy of GFlowNets (\ie, how states should move forward) while the reverse time SDE in \Eqref{eq:reverse_sde} models the backward policy of GFlowNets (\ie, how states should move backward).
In this case, a trajectory $\{\s_t\}_{t=0}^1$ is also in the form of  $\{\x_t\}_{t=0}^1$.
\end{observation}

Note that we cannot directly treat $\x_t$ to be a GFlowNet state, as the theory of GFlowNets requires the graph of all latent states to be a DAG (\ie, one cannot return to an already visited state).
This state augmenting operation~\citep{Bengio2021GFlowNetF}  induces the required DAGness, including in the SDE case. This follows because any $(\x, t) \to (\x', t')$ transition with $t \ge t'$ is forbidden.
Without loss of generality, we assume that the notation $\x_t$ as a GFlowNet state already contains the time stamp itself in the context below\footnote{This is equivalent to defining $\tilde\x_t = (\x_t, t)$ and conduct discussion with $\tilde\x_t$ instead.}.

We now point out an analogy between a stochastic processes property and a GFlowNet property:

\begin{observation}
The property of a stochastic process
\begin{equation}
\begin{aligned}
\label{eq:diffusion_markov}
&\int p(\x_{t-h}, t-h)\PP_h(\x_{t}|\x_{t-h}) 
\dif\x_{t-h} \\
& \quad = \int p(\x_t, t) \PP_h(\x_{t+h}|\x_t) \dif\x_{t+h} = p(\x_t, t),
\end{aligned}
\end{equation}
can be interpreted as the GFlowNets flow matching constraint
\begin{align}
\sum_{\s} F(\s, \s') = \sum_{\s{''}} F(\s', \s{''}) \triangleq F(\s'), \ \forall \s'\in\mathcal{S},
\end{align}
where we have $F(\s, \s') = F(\s)P_F(\s'|\s)$.
\end{observation}

We point out that \Eqref{eq:diffusion_markov} is a standard starting point for deriving the Fokker-Planck equation, as shown in Appendix:
\begin{proposition}[\citet{ksendal1985StochasticDE}]
\label{prop:fk_eq}
Taking the limit as $h \rightarrow 0$, \Eqref{eq:diffusion_markov} implies
\begin{equation}
\begin{split}
\partial_t p(\x, t)
= -\nabla_\x & \left(p(\x, t)\f(\x, t)\right) \\
& + \frac{1}{2}\nabla_\x^2\left(p(\x, t)g^2(\x, t)\right).
\end{split}
\end{equation}
\end{proposition}

\textbf{Equivalence between detailed balance and score matching.}
We investigate such a setting where
we  want to model the reverse process:
\begin{align}
\label{eq:modeling_rev_sde}
\dif \x &= \left[\f(\x)-g^2(t) \s(\x, t)\right]\dif \bar{t} + g(t)\dif \bar\w_t,   
\end{align}
where $\s(\cdot, \cdot):\R^D\times[0,1] \to \R^D$ is a neural network, and $\f(\cdot)$ and $g(\cdot)$ are given\footnote{$\f(\x), \s(\x)$ could also be written as $\f(\x_t, t), \s(\x_t, t)$ in a more strict / general way.}.
We propose to use the detailed balance criterion of GFlowNets  to learn this neural network.
From the above discussion, we can see there is an analogy realized by $F(\s_t)\approx p_t(\x)$.
We show the validity of such a strategy in the following proposition.

\begin{proposition}
\label{prop:db_sm}
GFlowNets' detailed balance condition
\begin{align}
\label{eq:db_sde}
\begin{split}
&\lim_{h \rightarrow 0} \frac{1}{\sqrt{h}}
\huge(\log p_t(\x_t) + \log P_F(\x_{t+h}|\x_t) -\\
& \quad \log p_{t+h}(\x_{t+h}) + \log P_B(\x_t|\x_{t+h})\huge)=0,
\end{split}
\end{align}
$(\forall\x_t\in\mathbb{R}^D, \forall t \in (0, 1))$ is equivalent to 
\begin{align*}
\eps^\top\left(\s(\x_{t},t) - \nabla_{\x}\log p_t(\x) \right) = 0, \forall\eps,\x_t\in\mathbb{R}^D, \forall t\in[0, 1],
\end{align*}
 which is the optimal solution to (sliced) score matching:
\begin{align*}
\min_{\s}\E_{\x\sim p_t}\E_\eps\left[\eps^\top\nabla_\x\s(\x)\eps + \frac{1}{2}\left(\eps^\top\s(\x, t)\right)^2\right], \forall t.
\end{align*}
\end{proposition}

\subsubsection{Schrödinger Bridge}

The Schr\"odinger Bridge or SB for short~\citep{schrodinger1932theorie,Leonard2013ASO,Chen2021OptimalTI} is a classical problem which solves the entropy regularized optimal transport.
In order to achieve this bridging target, the Iterative Proportional Fitting or IPF for short~\citep{Kullback1968ProbabilityDW} method proposes to solve the Schr\"odinger Bridge problem
with the following alternating optimization:
\begin{align}
    \label{eq:ipf_odd}
    \pi^{2m+1} &= \argmin\{\KL{\pi}{\pi^{2m}}: \pi_{\text{start}}=p_{\text{prior}}\},\\
    \label{eq:ipf_even}
    \pi^{2m+2} &= \argmin\{\KL{\pi}{\pi^{2m+1}}: \pi_{\text{end}}=p^{*}\},
\end{align}
where $\pi, \pi^m\in\mathcal{P}(\mathcal{X}^{n+1})$, namely distributions on $\mathcal{X}^{n+1}$ space, $\pi^0$ is initialized to some given base measure, and $\pi_{\text{start}},\pi_{\text{end}} \in \mathcal{P}(\mathcal{X}) $ refer to the marginal distribution of $\pi$ on the first and last index respectively. 

In practice, IPF models the joint distribution $\pi$ in \Eqref{eq:ipf_odd} in the decomposition form of forward probability product $\pi_{\text{start}}(\s_1)\prod_i p(\s_{i+1}|\s_i)$, as in  \Eqref{eq:hvae_decoder} and~\ref{eq:ddpm_pf}, as the marginal distribution on the first state is fixed. On the contrary, $\pi$ in \Eqref{eq:ipf_even} is modelled with backward probability decomposition $\pi_{\text{end}}(\s_n)\prod_i p(\s_{i}|\s_{i+1})$, as in \Eqref{eq:hvae_encoder} and~\ref{eq:ddpm_pb}. As a matter of fact, such a discrete-time SB formulation generalizes the DDPM by relaxing the constraint on its noise diffusion process. 
\begin{observation}
The discrete-time Schr\"odinger Bridge is a special case of GFlowNet. Compared to the DDPM formulation, SB does not use a fixed backward policy,
but learns both the forward and backward policies together.
\end{observation}

We remark that such alternating optimization is in the same spirit as the wake-sleep algorithm~\citep{Hinton1995TheA} for learning latent variable models without requiring something like the REINFORCE gradient estimator.
Regarding the continuous-time setup, a similar observation could be made that SB generalizes the ScoreSDE formulation by not using a fixed noising process. We refer to \citet{Bortoli2021DiffusionSB,Shi2022ConditionalSU} for practical guidance regarding learning diffusion SB for generative modeling.
\subsection{Exact Likelihood Models}

Autoregressive (AR) models can b e viewed as sampling $p(\x_{1:i+1}|\x_{1:i})$ sequentially one dimension $i$ at a time in order to generate the final vector $\x$.
\citet{Zhang2022GenerativeFN} use an AR-like (with a learnable ordering) model to parametrize the GFlowNet.
Indeed, we can define every (forward) action of the GFlowNet as specifying one more pixel on top of the current state, and the backward policy turns one pixel into an unspecified value~\citep{Zhang2022GenerativeFN}. This makes AR models
special cases of GFlowNet where the order in which the pixels are specified is fixed, making the GFlowNet DAG a tree \citep{Bengio2021FlowNB}.

\begin{observation}
The standard autoregressive model is a special kind of GFlowNet where
\begin{itemize}
    \item $\s_i := \x_{1:i}$ is the GFlowNet state;
    \item $P_F(\s_{i+1}|\s_i)=p(\x_{1:i+1}|\x_{1:i})$ is the forward policy;
    \item $P_B(\s_i|\s_{i+1})=
    \delta
    \{\s_i\ \textrm{comprises the first}\ i \text{ dimensions of}\ \s_{i+1}\}$,
\end{itemize}
where $\delta\{\cdot\}$ is the Dirac Delta distribution for continuous variables, and is the indicator function in the discrete case.
\end{observation}

This modeling makes the latent graph 
of the GFlowNet to be a tree; alternatively, if we allow a learnable ordering as with \citet{Zhang2022GenerativeFN}, the trajectories in latent space form a general DAG. This is related to non-autoregressive modeling methods in the NLP community \cite{Gu2018NonAutoregressiveNM}.

The normalizing flow or NF for short \citep{Dinh2015NICENI} is another way to sequentially construct desired data. It first samples $\z_1$ from a base distribution (usually the standard Gaussian), and then applies a series of invertible transformations $\z_{i+1}=\f(\z_i)$ until one finally obtains $\x:=\z_{n}$, where $n$ denotes the number of transformation layers.
\begin{observation}
The NF is a special kind of GFlowNet with deterministic forward and backward policies (except the first transition step), and $\s_i = \z_i$ are GFlowNet states.
\end{observation}
With the GFlowNet implementation of NF, the base distribution is the first (and only stochastic) step, while the other steps are deterministic. When both $P_F(\z_{i+1}|\z_i)$ and $P_B(\z_i|\z_{i+1})$ are deterministic (each being a Dirac at the value of some function applied to the conditioning argument) and match each other, it must be that they correspond to invertible functions. 
We next discuss maximum likelihood estimation (MLE) of AR and NF models.

\textbf{About MLE training.}
AR models and NFs are usually trained with MLE.
Although the likelihood of general GFlowNets is intractable, we lower bound it:
\begin{align}
\label{eq:log_pT_integral}
\log p_T(\x) &= \int_{\x\in\tau} P_F(\tau)\dif\tau \\
&= \log \E_{P_B(\tau|\x)}\left[\frac{P_F(\tau)}{P_B(\tau|\x)}\right] \\
&\ge \E_{P_B(\tau|\x)}\left[\log\frac{P_F(\tau)}{P_B(\tau|\x)}\right] \\
&= -\KL{P_B(\tau|\x)}{P_F(\tau)}, \label{eq:kl_lb}
\end{align}
This could again correspond to a kind of trajectory balance objective (KL-trajectory balance) as in Proposition~\ref{prop:hvae_kl_elbo}. Notice that this derivation is applicable to all GFlowNet specifications, rather than just exact likelihood models.
An IWAE-type bound~\citep{Burda2016ImportanceWA} is also applicable.
When we are given data points $\x$, we can directly use $\log P_B(\tau|\x) - \log P_F(\tau)$ as a sample-based tractable training loss for $\x\in \D$, with $\tau\sim P_B(\tau|\x)$, to maximize a variational lower bound on the log-likelihood.
Furthermore, if we $P_B(\tau|\x)$ is deterministic, this corresponds to a fixed ordering (a single trajectory) to construct $\x$, which is the  AR interpretation of a GFlowNet, and minimizing $\log P_B(\tau|\x) - \log P_F(\tau)$ is the same thing as maximizing the likelihood $\log P_F(\tau)$ with $\tau$ corresponding to $\x$, i.e., the MLE loss of AR models.

Summarizing, both AR models and NFs are GFlowNets with a tree-structured latent state graph, making every terminating state reachable  by only one trajectory.

\subsection{Learning a Reward Function from Data}

Approximately sampling from an energy-based model (EBM) can be obtained from a GFlowNet whose negative log of the reward function is the energy function. 
We could use any GFlowNet model including those discussed in previous sections, and jointly train it together with the EBM.
For instance, in the EB-GFN  \cite{Zhang2022GenerativeFN} algorithm a GFlowNet is used to amortize the computational MCMC process of the EBM contrastive divergence training.
The two models (EBM and GFlowNet) are updated alternately.

The GAN \citep{Goodfellow2014GenerativeAN} is closely related to EBMs \cite{Che2020YourGI}, while its algorithm is more computationally efficient. However, though it may look reasonable at first glance, we cannot directly use the discriminator $D(\x)$ as the reward for GFlowNet training. If we did, at the end of perfect training, we would get an optimal discriminator $D^*(\x)=\frac{p^*(\x)}{p^*(\x) + p_T(\x)}$, 
and the optimized GFlowNet terminating distribution would be $p_T(\x)\propto D^*(\x)$. This cannot induce $p_T(\x)=p^*(\x)$. In fact, if $p_T(\x)=p^*(\x)$, we will have $D^*(\x)\equiv 1/2$ and $p_T(\x)=p^*(\x)\equiv \text{constant}$, which is impossible for general data with unbounded support.
To fill this gap, we could instead use the following algorithm.

\begin{proposition}
\label{prop:gfn_gan}
An alternative algorithm which trains the discriminator to distinguish between generated data and true data, and trains the GFlowNet with negative energy
\begin{equation}
\log \frac{D(\x)}{1-D(\x)}  + \log p_T(\x)
\end{equation}
would result in a valid generative model.
\end{proposition}

Nonetheless, we unfortunately do not have access to the exact value of $p_T(\x)$ if the generator is a general GFlowNet (\Eqref{eq:log_pT_integral}), which makes this algorithm intractable.

\section{Generative Modeling and Sampling}

\label{sec:gm_sampling}

Despite all the connections described above, the similarity between GFlowNets and generative models mainly exists on the modeling side. As a matter of fact, GFlowNets in its origin are designed for sampling, which is a very different problem setup from generative modeling. In generative modeling, a training dataset is provided and is treated as an empirical approximation of the target distribution. However, in sampling problems where GFlowNets are proposed to be used, the practitioners are given a black-box (probably unnormalized) target density function $p^*(\cdot)$ as a callable oracle instead of a dataset. That means that in sampling, the learning signal comes from a non-differentiable function which takes in a data point $\x$ and returns a scalar $p^*(\x)$. In theory, the probability density function of the target distribution contains all the information about the distribution, and it would be possible to use MCMC sampling to draw a dataset from this density function. Nonetheless, in practice it is computationally infeasible to explore all the modes of its landscape (due to the high dimensionality), especially when the distribution is not unimodal, let alone the MCMC algorithms would take infinite computation time to mix. From this perspective, both generative modeling and sampling are aiming at learning probabilistic models to represent some target distributions, but the former is an easier distribution matching problem than the latter: in generative modeling, the exploration part of sampling has been done, and we only need to exploit the information in the dataset by fitting the generative models. 

Under the context of reinforcement learning, the connection between generative modeling and sampling is similar to the relationship between offline RL~\citep{Lin2004SelfImprovingRA,Lange2012BatchRL} and online RL, where in offline RL we are given a dataset of labelled trajectories obtained from the interaction between a predefined expert agent and a particular environment.
It is well known that in online RL, due to the high complexity of the settings, algorithms would give results with high variance and large stochasticity~\citep{Henderson2017DeepRL}. Besides, works have shown that the performance of offline RL tasks are much more stable than their online variants~\citep{Agarwal2019StrivingFS}.
This originates from the fact that in online RL the agent need to interact with the environment to explore the landscape of the task, which is full of uncertainty. The exploration is also a serious challenge that sampling would face.
On the other hand, in offline RL the goal is simply to learn an optimal policy from existing data, which is related to imitation learning -- generative modeling in the trajectory level.

\section{Towards Improved Generative Modeling with Insights from GFlowNets}
\label{sec:mle_gfn_alg}
We have provided a GFlowNet-based probabilistic framework to unify the generative behaviors of different classes of generative models. 
In this section, we investigate that whether we could further boost the performance of generative modeling with insights from GFlowNets.

\begin{algorithm}[t]
\caption{Generative modeling via
maximum likelihood estimation based
GFlowNet (MLE-GFN) training}
\label{alg:tbc_training}
\begin{algorithmic}[1]
\REQUIRE 
Training dataset $\D=\{\x_i\}_i$,
GFlowNet policy $P_F(\tau;\vtheta), P_B(\tau|\x;\vphi)$ with parameters $\vtheta, \vphi$. 
\REPEAT
\STATE Uniformly sample $\x$ from dataset $\D$;
\STATE $\triangle\vtheta \gets \nabla_{\vtheta}\KL{P_B(\cdot|\x;\vphi)}{P_F(\cdot;\vtheta)}$;
\STATE Sample backward trajectories $\tau,\tau'\sim P_B(\cdot|\x;\vphi)$;
\STATE $\triangle\vphi\gets \nabla_{\vphi}\ltbc(\tau,\tau')$;
\STATE Update $\vtheta,\vphi$ with some optimizer;
\UNTIL 
some convergence condition
\end{algorithmic}
\end{algorithm}

\begin{table*}[t]
\setlength{\tabcolsep}{2.5mm}
\centering
\caption{
Experiment results with seven 2D synthetic problems.
We display the MMD (in units of $1\times 10^{-4}$).
}
\label{tab:synthetic_mmd}
\begin{tabular}{c|l|ccccccc}
\toprule
Metric & Method & 2spirals & 8gaussians & circles & moons & pinwheel & swissroll & checkerboard \\
\midrule
\multirow{4}{*}{MMD$\downarrow$} 
& PCD    &  $2.160$&$0.954$&$0.188$&$0.962$&$0.505$&$1.382$& $2.831$ \\
& ALOE   &  $21.926$ & $107.320$ & $0.497$ & $26.894$ & $39.091$ & $0.471$ & $61.562$ \\
& ALOE+  &  $ \textbf{0.149} $ & ${0.078}$ & $0.636$ & $0.516$ & $1.746$ & $0.718$ & $12.138$ \\
& EB-GFN &  $0.583$ & $0.531$ & ${0.305}$ & ${0.121}$ & ${0.492}$ & ${0.274}$ & ${1.206}$\\ 
& MLE-GFN 
& ${0.472}$ & $\textbf{0.046}$ & $\textbf{-0.026}$ & $\textbf{-0.021}$ & $\textbf{0.211}$ & $\textbf{0.151}$ & $\textbf{0.393}$\\  
\bottomrule
\end{tabular}

\end{table*}

\subsection{Trajectory Balance Consistency}

Since the gap between the inherent difference between generative modeling with GFlowNet sampling discussed in \Secref{sec:gm_sampling}, there is no straightforward way to combine algorithms from these two worlds. To see this, recall that in the original GFlowNet trajectory balance objective, the reward value $R(\x)$ is needed:
\begin{align}
    \mathcal{L}_{\text{TB}}(\tau) &= \left[\log\frac{Z P_F(\tau)}{R(\x) P_B(\tau|\x)}\right]^2,
\end{align}
where $Z$ is a learnable scalar parameter, $\tau=(\s_1, \ldots,\s_{n})$ and $\s_{n}=\x$.
However, this is not practicable since in generative modeling we do not have access to the callable target density function $p^*(\x) \propto R(\x)$. To circumvent this obstacle, notice that if a GFlowNet with infinite capacity is trained to completion, we would have
\begin{align}
    \frac{P_F(\tau)}{P_B(\tau|\x)} = \frac{R(\x)}{Z} = \frac{P_F(\tau')}{P_B(\tau'|\x)}
\end{align}
for any two different trajectories $\tau, \tau'$ with the same terminating state $\x$. Consequently, we propose such consistency objective to avoid the appearance of the reward term,
\begin{align}
\label{eq:tbc}
    \ltbc(\tau,\tau') = \left[\log\frac{P_F(\tau)}{P_B(\tau|\x)}- \log\frac{P_F(\tau')}{P_B(\tau'|\x)}\right]^2.
\end{align}
Here we use ``TBC" to denote ``trajectory balance consistency".
The proposed consistency loss objective only assures the balance between the forward and backward trajectories of GFlowNet model, but receives no signal about information of the target distribution that the GFlowNet desire to match. Hence we cannot use $\ltbc$ as the only training loss even with $\x$ taken from the training set, which would potentially end up with a naive solution, \eg, constant output GFlowNet policy. Howbeit, we propose to combine the trajectory balance consistency as a regularization technique with the original generative modeling methods. We would demonstrate the regularization efficacy of the proposed strategy in the following section. 

Following the analysis from \citet{Malkin2022GFlowNetsAV}, we show the relationship between the proposed GFlowNet balance objective and divergence minimizing objective.
\begin{proposition}
\label{prop:tbc_kl}
    Denote the parameters of the backward policy $P_B$ by $\vphi$, then the gradient of the $\ltbc$ objective defined in~\Eqref{eq:tbc} with respect to $\vphi$ satisfies
    \begin{align}
        \frac{1}{4}\E_{\tau,\tau'\sim P_B}\left[\nabla_{\vphi}\ltbc(\tau,\tau')\right] = \nabla_{\vphi} \KL{P_B}{P_F}.
    \end{align}
\end{proposition}

The proposition demonstrates the correctness of optimizing on $\ltbc$ to minimize the divergence.
This indicates that, as a general rule of generative modeling, we could optimize the forward policy with the variational bound in \Eqref{eq:kl_lb}, and optimize the backward policy with $\ltbc$ defined in \Eqref{eq:tbc}. We specify our method in Algorithm~\ref{alg:tbc_training}.
We refer to the algorithm as MLE-GFN since it is essentially optimizing a variational bound of the model likelihood.
\subsection{Synthetic Demonstration}

The experiment in this subsection follows the setup from ~\citet{Dai2020LearningDE,Zhang2022GenerativeFN}. The objective is to model $7$ different distribution over $32$-dimensional binary space, as displayed in \Figref{fig:gfn_synthetic_result}. The binary data is quantized from $2$ dimensional continuous data via the Gray code~\citep{gray1953pulse}. We consider with the algorithm and baselines from \citet{Zhang2022GenerativeFN}, including persistent contrastive divergence or PCD for short \citep{Tieleman2008TrainingRB}, ALOE~\citep{Dai2020LearningDE}, and energy-based GFlowNet (EB-GFN). 

For the proposed trajectory balance consistency augmented GFlowNet (MLE-GFN) method, we follow the same GFlowNet modeling as in EB-GFN, just using the novel objectives in Algorithm~\ref{alg:tbc_training} to separately learn the forward and backward policies. Notice that EB-GFN needs to learn an additional energy function, thus consumes larger number of parameters. In the result presentation, ``ALOE" and ``ALOE+" denote two different modeling methods, where the former shares a similar number of parameters to the GFlowNets, while the latter is thirty times larger.
For evaluation, we report the MMD~\citep{Gretton2012AKT} between ground truth samples and generated samples. 
We demonstrate quantitative results in Table~\ref{tab:synthetic_mmd}, where the proposed algorithm outperform most other baselines. 
We defer other results and details to \Secref{sec:synthetic_appendix}.

\subsection{DDPM Demonstration}

\begin{table}[t]

    \caption{Comparison between the proposed MLE-GFN in Algorithm~\ref{alg:tbc_training} and the baseline on CIFAR-$10$. We evaluate both the sample quality (FID) and likelihood (NLL).
    We train models with a smaller level of computation resource, which explains the performance gap with the results in original DDPM paper.
    }
    \label{tab:ddpm_result}
    
    \centering
    \begin{tabular}{l|cc}
    \toprule
      Method  & FID$\downarrow$ & NLL$\downarrow$ \\
    \midrule
       Baseline & $10.616$ & $4.276$ \\
       iDDPM &  $10.823$ & $4.198$  \\
       MLE-GFN & $\textbf{10.062}$ & $\textbf{4.157}$  \\
    \bottomrule
    \end{tabular}

\end{table}

DDPM~\citep{Ho2020DenoisingDP} defines a diffusion generative process and has achieved great success in high quality visual generation. In \Secref{sec:diffusion}, we have pointed out their similarity on modeling behavior: when we parameterize the forward and backward policy with particular Gaussian distributions as in Observation~\ref{obs:ddpm}, a GFlowNet is crystallized into a DDPM. The forward policy 
$P_F$ is the denoised reverse process $p$, while the backward policy $P_B$ is the diffusion process $q$ which involves noises.

DDPM use a fixed $P_B$ (and hence fixed $\{\beta_i\}_i$) to enable a stable training process\footnote{It is mentioned in Section 4.2 of \citet{Ho2020DenoisingDP} that ``incorporating a parameterized variance leads to unstable training and poor sample quality".}. In this section, we propose to utilize the proposed consistency objective to ``defreeze" the constant for better expressiveness and sample quality, following the guideline in Algorithm~\ref{alg:tbc_training}. Concretely, under the context of DDPM modeling, the likelihood variational lower bound reduces to the original DDPM denoising objective (see Proposition~\ref{prop:ddpm_elbo}), and the consistency objective is used to update the variance parameter $\{\beta_i\}_i$.
Notice that there are other works, \eg, \citet{Nichol2021ImprovedDD}, that also propose to achieve a similar goal, but from different angles. In this section, we are not aiming for a state-of-the-art performance, but mainly for a demonstration of the application of GFlowNet idea upon generative modeling tasks.

Due to computation resource limitation, we train a DDPM-specified GFlowNet on the CIFAR-$10$ dataset with $n=100$ (\ie, GFlowNet trajectory length) denoising steps for $200$k training steps. This is less then the $n=1000$ and $800$k step training by the original diffusion.
We defer other training and experimental details to \Secref{sec:app_experiment}. 
We first directly quantitatively show the sample quality and likelihood evaluation metric of the proposed GFlowNet-inspired consistency augmented algorithm and the DDPM baseline in Table~\ref{tab:ddpm_result}. Here the sample quality is measured by the FID score~\citep{Heusel2017GANsTB}. Apart from the DDPM baseline, we also compare with the hybrid objective proposed in improved DDPM~\citep[iDDPM]{Nichol2021ImprovedDD}, which is designed to improve the likelihood evaluation metric. iDDPM will actually achieve worse sample quality while having better likelihood; on the other hand, our proposed approach could achieve a better trade-off, achieving both better likelihood and FID performance than the baseline.

\begin{figure}[t]
    \centering
    \includegraphics[width=0.99\linewidth]{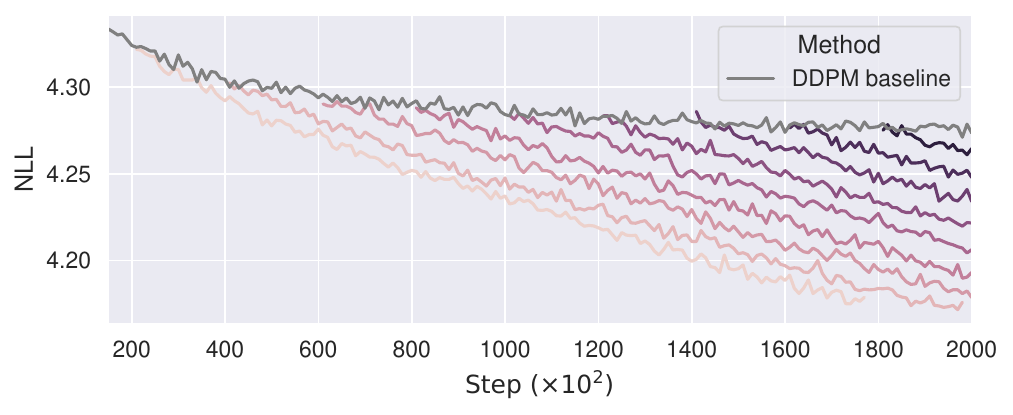}
    
    \caption{NLL performance of baseline and proposed methods. 
    The gray line shows the results of the DDPM baseline; the colored lines show the  results of the proposed GFlowNet consistency augmented training started with the pretrained weights of the baseline method from $20000, 40000, \ldots, 180000$ steps, respectively.}
    \label{fig:ltbc_curve_nll}
    
\end{figure}

What's more, we also try to finetune a pretrained model with our proposed method. Specifically, we train a DDPM model with a fixed number of steps, and then switch to the proposed MLE-GFN to continue the training. \Figref{fig:ltbc_curve_nll} presents the curves of likelihood evaluation
for the DDPM baseline (grey curve) and the proposed method (violet curves).

\section{Conclusion}

We have interpreted existing generative models as GFlowNets with different policies over sample trajectories. This provides some insight into the overlaps between existing generative modeling frameworks, and their connection to general-purpose algorithms for training them.
Furthermore, this unification implies a method of constructing an agglomeration of varying types of generative modeling approaches, where GFlowNet acts as a general-purpose glue for tractable inference and training.

\section*{Acknowledgments}

D.Z., N.M., and Y.B.\ acknowledge funding from CIFAR, Genentech, Samsung, and IBM.

N.M.\ and D.Z. thank Xu Ji and L\'ena N\'ehale Ezzine for helpful discussions that led to the development of the TBC objective.

\newpage
\bibliography{ref}
\bibliographystyle{icml2023}

\newpage

\onecolumn

\appendix

\section{Summary of Notation}
\begin{table}[h]
\centering
\begin{tabular}{ll}
    \toprule
    Symbol & Description \\ %
    \midrule
    $\mathcal{S}$ & GFlowNet state space \\
    $\mathcal{X}$ & object (terminal state) space, subset of $\mathcal{S}$\\
    $\mathcal{A}$ & action / transition space (edges $\s\to\s'$)\\
    $\mathcal{G}$ & directed acyclic graph $(\mathcal{S},\mathcal{A})$ \\
    $\mathcal{T}$ & set of complete trajectories \\
    $\s$ & state in $\mathcal{S}$\\
    $\s_0$ & initial state, element of $\mathcal{S}$ \\
    $\x$ & terminal state in $\mathcal{X}$ \\
    $\tau$ & trajectory in $\mathcal{T}$ \\
    $F: \mathcal{T} \to \R$ & Markovian flow\\
    $F: \mathcal{S} \to \R$ & state flow\\
    $F: \mathcal{A} \to \R$ & edge flow\\
    $P_F(\s'|\s)$ & forward policy (distribution over children) \\
    $P_B(\s|\s')$ & backward policy (distribution over parents) \\
    $P_T(\x)$ & terminating distribution \\
    $Z$ & scalar, equal to $\sum_{\tau\in\mathcal{T}}F(\tau)$ for a Markovian flow \\
    \bottomrule
\end{tabular}
\end{table}

\section{Proofs}
\label{sec:proofs}

\subsection{Proposition~\ref{prop:hvae_kl_elbo}}
\label{sec:proof_hvae_elbo}
\begin{proof}
We have
\begin{align*}
\KL{ P_B(\tau)}{P_F(\tau)} & =\E_{p^*(\x)q(\z_{1:n-1}|\x)}\left[\log\frac{p^*(\x)q(\z_{1:n-1}|\x)}{p(\z_{1:n-1})p(\x|\z_{1:n-1})}\right] \\
&= -\E_{p^*(\x)} \E_{q(\z_{1:n-1}|\x)}\left[\log\frac{p(\x,\z_{1:n-1})}{q(\z_{1:n-1}|\x)}\right] + \E_{p^*(\x)}\left[\log p^*(\x)\right]\\
&= -\E_{p^*(\x)}\left[\textrm{ELBO}_{p, q}(\x)\right] - \mathcal{H}[p^*(\cdot)].
\end{align*}
Here $\mathcal{H}[p^*(\cdot)]$ denotes the entropy of the target distribution, which is a constant w.r.t. GFlowNet parameters.
\end{proof}

\subsection{Proposition~\ref{prop:ddpm_elbo}}
We follow a similar derivation of \citet{Ho2020DenoisingDP} and \Secref{sec:proof_hvae_elbo}.
\begin{align*}
\KL{ P_B(\tau)}{P_F(\tau)} &\cong -\E_{p^*(\x)}\left[\textrm{ELBO}_{p, q}(\x)\right] \\
\textrm{ELBO}_{p, q}(\x) &= \E_q\left[-\log \frac{p(\z_{1:n})}{q(\z_{1:n-1}| \x)}\right] =  \E_q\left[-\log p(\z_1) -\sum_{i<L}\log \frac{p(\z_{i+1}|\z_i)}{q(\z_{i}|\z_{i+1})} - \log\frac{p(\z_{n}|\z_L)}{q(\z_L|\z_{n})}\right]\\
&= \E_q\left[-\log \frac{p(\z_1)}{q(\z_1|\z_{n})} -\sum_{i<L}\log \frac{p(\z_{i+1}|\z_i)}{q(\z_{i+1}|\z_{i}, \z_{n})} - \log p(\z_{n}|\z_L)\right]\\
&\cong \E_q\left[\sum_{i<L}\KL{q(\z_{i+1}|\z_{i}, \z_{n})}{p(\z_{i+1}|\z_i)} - \log p(\z_{n}|\z_L)\right]
\end{align*}
Since the KL divergence between two Gaussian distributions has a close form, we have 
\begin{align*}
    \KL{q(\z_{i+1}|\z_{i}, \z_{n})}{p(\z_{i+1}|\z_i)} &\cong C_i\cdot \E\left[\left\Vert\boldsymbol{\mu}_i(\z_i) - \tilde{\boldsymbol{\mu}}_i(\z_i)\right\Vert^2\right], \\
    \E_q\left[-\log p(\z_{n}|\z_L)\right] &\cong C_L\cdot \E\left[\left\Vert\boldsymbol{\mu}_L(\z_i) - \z_{n}\right\Vert^2\right],
\end{align*}
where $\{C_i\}_i$ are scalar constants, and $\tilde{\boldsymbol{\mu}}_i(\z_i)$ is the mean value of $q(\z_{i+1}|\z_{i}, \z_{n})$, and $\tilde{\boldsymbol{\mu}}_i(\z) = \tilde C_i^1\z+\tilde C_i^2\x $ is a fixed (\ie, non-learnable) transformation defined with $\{\beta_i\}_i$ and $\x=\z_{n}$. Here $\tilde C_i^1, \tilde C_i^2$ are deterministic functions of $\{\beta_i\}_i$ defined by \citet{Ho2020DenoisingDP}. This derivation indicates that training a GFlowNet with KL-trajectory balance objective would end up with an effectively same (\ie, ignoring the constants) objective with the regression loss proposed in denoised autoencdoer~\citep{Vincent2008ExtractingAC} and denoising diffusion probabilistic models.

\subsection{Proposition~\ref{prop:fk_eq}}
\begin{proof}
First notice
\begin{align*}
\partial_t p(\x, t) \triangleq \lim_{h\to 0}\frac{1}{h}\left(p(\x, t+h)-p(\x, t)\right)
= \lim_{h\to 0}\frac{1}{h}\left(\int p(\x', t)\PP_h(\x_{}|\x')\dif\x' -p(\x, t)\right).
\end{align*}

Then for any function $w(\x)$, we have

\begin{align*}
&\quad \int w(\x) \partial_t p(\x, t) \dif\x \\
&= \int w(\x)\lim_{h\to 0}\frac{1}{h}\left(\int p(\x', t)\PP_h(\x_{}|\x')\dif\x' -p(\x, t)\right)\dif\x\\
&= \lim_{h\to 0}\frac{1}{h}\Large(\int w(\x)\int p(\x', t)\PP_h(\x_{}|\x')\dif\x'\dif\x -\int w(\x') p(\x', t) \int\PP_h(\x|\x')\dif\x \dif\x'\Large) \\
&= \lim_{h\to 0}\frac{1}{h}\int p(\x', t)\PP_h(\x|\x')\left(w(\x) - w(\x')\right)\dif\x\dif\x'\\
&\triangleq\int p(\x', t) \sum_{n=1} w^{(n)}(\x')D_n(\x') \x' \\
&= \int w(\x')\sum_{n=1}\left(-\frac{\partial}{\partial\x'}\right)^n\left( p(\x',t)D_n(\x')\right)\dif\x',
\end{align*}

where $D_n(\x')=\lim_{h\to 0}\frac{1}{h n!}\int \PP_h(\x|\x') (\x-\x')^n\dif\x$ and the last step uses integral by parts.
This tells us 
\begin{align*}
\partial_t p(\x, t) = \sum_{n=1}\left(-\frac{\partial}{\partial\x}\right)^n\left( p(\x,t)D_n(\x)\right)
= -\nabla_\x \left(p(\x, t)\f(\x, t)\right) + \frac{1}{2}\nabla_\x^2\left(p(\x, t)g^2(\x, t)\right),
\end{align*}
which is essentially the Fokker-Planck equation.

\end{proof}

\subsection{Proposition~\ref{prop:db_sm}}
From the above SDEs, we know that the forward and backward policy is
\begin{align*}
\x_{t+h} &= \x_t + \f(\x_t)h + \sqrt{h}g(t)\cdot \dd_F, \\
\x_t &= \x_{t+h} + \left[g^2(t+h)\s(\x_{t+h}) - \f(\x_{t+h})\right]h + \sqrt{h}g(t+h)\dd_B,
\end{align*}
where $\dd_F, \dd_B\sim \N(0, \mathbf{I}_D)$.
Since we know $h \to 0$, the left and right side of ~\Eqref{eq:db_sde} become
\begin{align*}
\log P_F(\x_{t+h}|\x_t) -\log P_B(\x_t|\x_{t+h}) &= -\frac{D}{2}\log(2\pi g^2_t h) - \frac{1}{2g_t^2 h}\Vert \Delta\x -h\f_t \Vert^2 \\
& + \frac{D}{2}\log(2\pi g^2_{t+h} h) +  \frac{1}{2g_{t+h}^2 h}\Vert \Delta\x -h\f_{t+h} + hg^2_{t+h}\s(\x_{t+h}) \Vert^2,\\
\log p_{t+h}(\x_{t+h}) - \log p_t(\x_t) &= \Delta\x^\top\nabla_{\x}\log p_t(\x) + O(h),
\end{align*}
where $g_t=g(t), \f_t = \f(\x_t), \Delta\x = \x_{t+h} -\x_t = \sqrt{h}g_t\eps, \eps\sim\N(0, \mathbf{I}_D)$. Therefore,

\begin{align*}
\lim_{h\to 0}\frac{1}{\sqrt{h}g_t}(\log & p_{t+h}(\x_{t+h})x - \log p_t(\x_t)) = \eps^\top \nabla_{\x}\log p_t(\x),\\
\lim_{h\to 0} \frac{1}{\sqrt{h}g_t}(\log& P_F(\x_{t+h}|\x_t) -\log P_B(\x_t|\x_{t+h})) \\
&=\lim_{h\to 0} \frac{D}{\sqrt{h}g_t}(\log g_{t+h}- \log g_t) 
-\frac{1}{2g_t^3h^{3/2}}\left(h g_t^2\Vert\eps\Vert^2 + h^2\Vert\f_t\Vert^2 -2h^{3/2}g_t\eps^\top\f_t\right) \\
&\phantom{aaaaaaaaaaaaa} + \frac{1}{2g_tg_{t+h}^2h^{3/2}}\left(hg_t^2\Vert\eps\Vert^2 - 2h^{3/2}g_t\eps^\top\f_{t+h} + 2g_tg_{t+h}^2h^{3/2}\eps^\top\s(\x_{t+h})\right) \\
&= \lim_{h\to 0} \eps^\top\left(\frac{\f_t}{g_t^2} - \frac{\f_{t+h}}{g_{t+h}^2}\right) + \eps^T\s(\x_{t+h}) = \eps^T\s(\x_{t}),
\end{align*}

with some smoothness assumptions on $g(t)$ and $\f(\x_{t})/g(t)$.
This tells us that in order to satisfy the detailed balance criterion, we need to satisfy
\begin{align*}
    \eps^\top\left(\s(\x_{t},t) - \nabla_{\x}\log p_t(\x) \right) = 0, \forall\eps,\x_t\in\mathbb{R}^D, \forall t\in[0, 1].
\end{align*}

Since $\eps$ could take any value, this is equivalent that the model $\s$ should match with score function $\nabla_\x \log p(\x)$.
Also, note that this is exactly sliced score matching \cite{Song2019SlicedSM}, which has a more practical formulation
\begin{align*}
\E_\eps\E_{\x\sim p}\left[\eps^\top\nabla_\x\s(\x)\eps + \frac{1}{2}\left(\eps^\top\s(\x)\right)^2\right].
\end{align*}

\subsection{Proposition~\ref{prop:gfn_gan}}
When both models (the GFlowNet and the discriminator) are trained perfectly, we have
\begin{align*}
D^*(\x) = \frac{p^*(\x)}{p^*(\x) + p_T(\x)},
\end{align*}    
and thus 
\begin{align*}
p_T(\x) \propto \exp\left(\log \frac{D^*(\x)}{1-D^*(\x)}  + \log p_T(\x)\right) = \exp\left(\log \frac{p^*(\x)}{p_T(\x)} + \log p_T(\x)\right) = p^*(\x).
\end{align*}
Hence it is a valid generative model algorithm.

\subsection{Proposition~\ref{prop:tbc_kl}}
Because
$$
\nabla_{\vphi}\ltbc(\tau,\tau') = 2\left(\log\frac{P_F(\tau)}{P_B(\tau|\x;\vphi)} - \log\frac{P_F(\tau')}{P_B(\tau'|\x;\vphi)}\right)\left(-\nabla_{\vphi}\log P_B(\tau|\x;\vphi) + \nabla_{\vphi}\log P_B(\tau'|\x;\vphi)\right),
$$
we have
\begin{align*}
    \frac{1}{2}\E_{\tau,\tau'\sim P_B}&\left[\nabla_{\vphi}\ltbc(\tau,\tau')\right] \\
    =& -\E_{\tau\sim P_B}\left[\log\frac{P_F(\tau)}{P_B(\tau|\x;\vphi)}\nabla_{\vphi}\log P_B(\tau|\x;\vphi)\right] + \E_{\tau,\tau'\sim P_B}\left[\log\frac{P_F(\tau)}{P_B(\tau|\x;\vphi)}\nabla_{\vphi}\log P_B(\tau'|\x;\vphi)\right] \\
    & -\E_{\tau'\sim P_B}\left[\log\frac{P_F(\tau')}{P_B(\tau'|\x;\vphi)}\nabla_{\vphi}\log P_B(\tau'|\x;\vphi)\right] + \E_{\tau,\tau'\sim P_B}\left[\log\frac{P_F(\tau')}{P_B(\tau'|\x;\vphi)}\nabla_{\vphi}\log P_B(\tau|\x;\vphi)\right] \\
    =& -2\E_{\tau\sim P_B}\left[\log\frac{P_F(\tau)}{P_B(\tau|\x;\vphi)}\nabla_{\vphi}\log P_B(\tau|\x;\vphi)\right] + \E_{\tau\sim P_B}\left[\log\frac{P_F(\tau)}{P_B(\tau|\x;\vphi)}\right]\underbrace{\E_{\tau'\sim P_B}\left[\nabla_{\vphi}\log P_B(\tau'|\x;\vphi)\right]}_{=0} \\
    =& 2\nabla_{\vphi}\E_{\tau\sim P_B}\left[\log\frac{P_B(\tau|\x;\vphi)}{P_F(\tau)}\right] \\
    =& 2\nabla_{\vphi}\KL{P_B}{P_F}
\end{align*}

\section{More about Experimental Demonstration}
\label{sec:app_experiment}

\subsection{Synthetic Demonstration}
\label{sec:synthetic_appendix}

\newcommand\mpwid{0.16}
\newcommand\hinterval{0.5cm}
\newcommand\samplempwid{0.146}
\newcommand\samplehinterval{0.23cm}
\newcommand\samplefigwid{\textwidth}
\begin{figure*}[t]
\centering
\begin{minipage}{\textwidth}
    \begin{minipage}[t]{\samplempwid\textwidth}
    \centering
    \small{2spirals}\\
    \includegraphics[width=\samplefigwid,
    ]{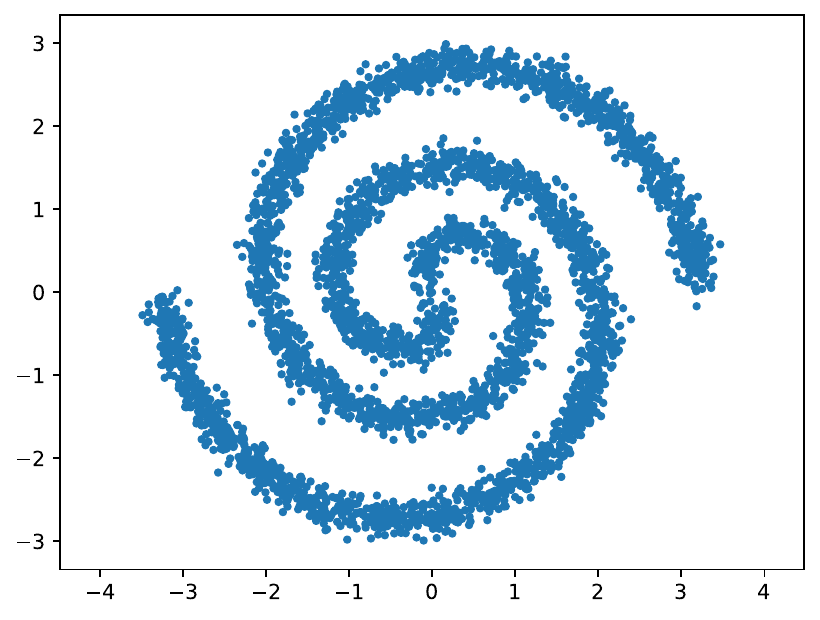}
    \end{minipage}
\hspace{-\samplehinterval}
    \begin{minipage}[t]{\samplempwid\textwidth}
    \centering
    \small{8gaussians}\\
    \includegraphics[width=\samplefigwid
    ]{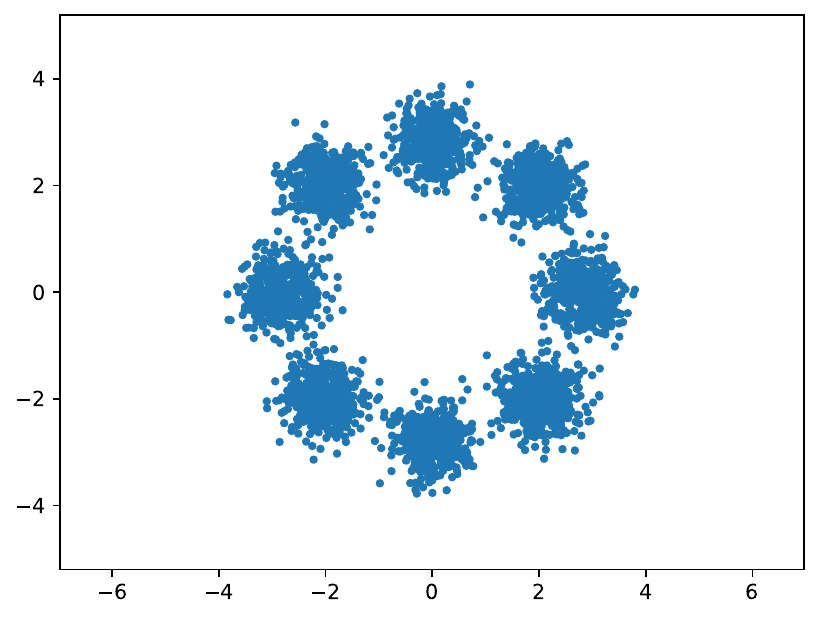}
    \end{minipage}
\hspace{-\samplehinterval}
    \begin{minipage}[t]{\samplempwid\textwidth}
    \centering
    \small{circles}\\
    \includegraphics[width=\samplefigwid
    ]{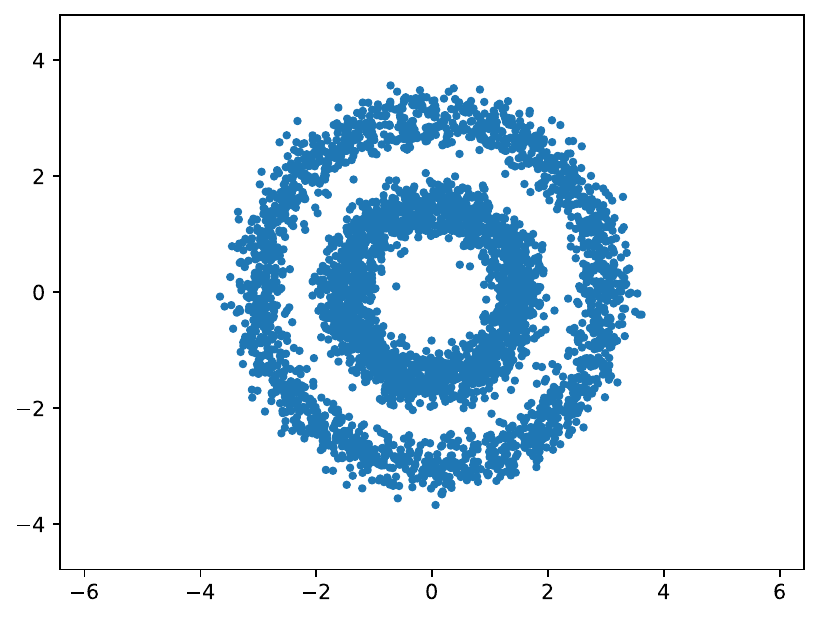}
    \end{minipage}
\hspace{-\samplehinterval}
    \begin{minipage}[t]{\samplempwid\textwidth}
    \centering
    \small{moons}\\
    \includegraphics[width=\samplefigwid
    ]{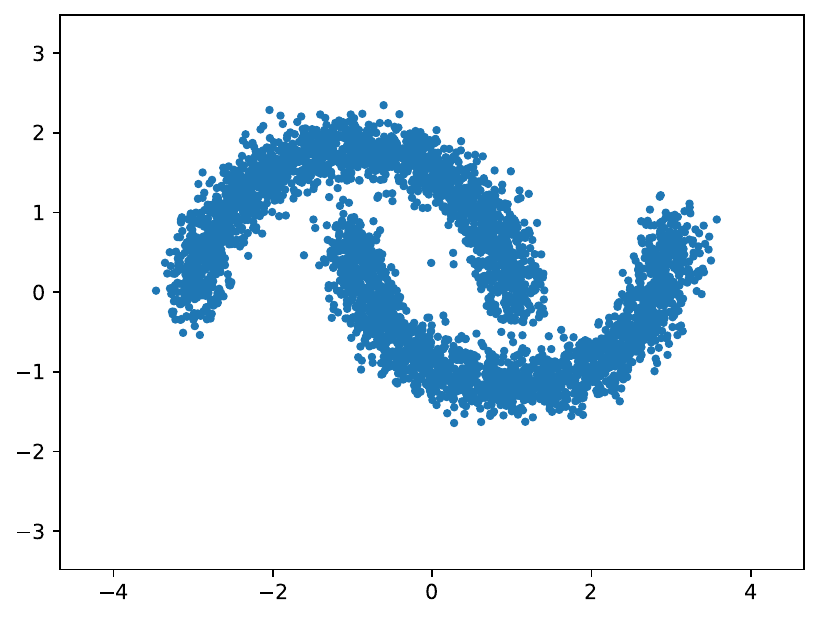}
    \end{minipage}
\hspace{-\samplehinterval}
    \begin{minipage}[t]{\samplempwid\textwidth}
    \centering
    \small{pinwheel}\\
    \includegraphics[width=\samplefigwid
    ]{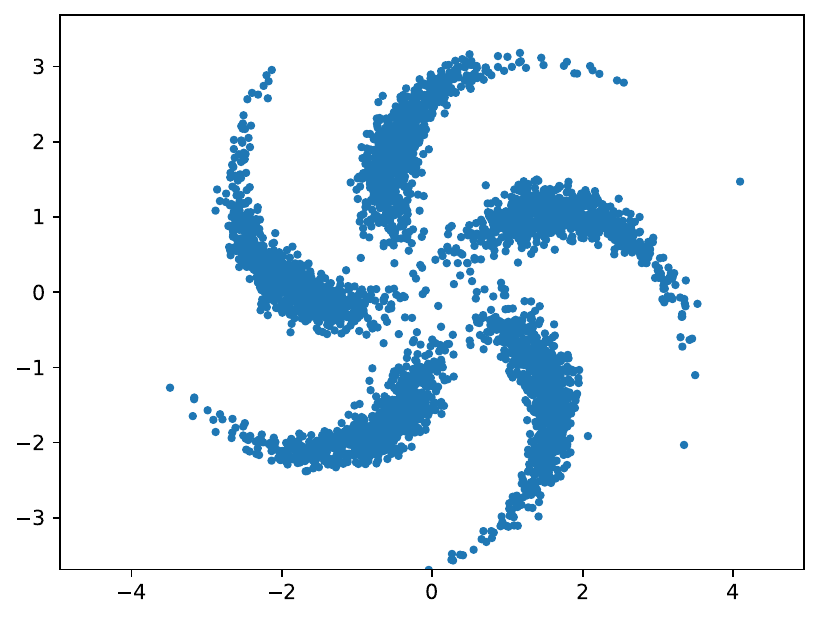}
    \end{minipage}
\hspace{-\samplehinterval}
    \begin{minipage}[t]{\samplempwid\textwidth}
    \centering
    \small{swissroll}\\
    \includegraphics[width=\samplefigwid
    ]{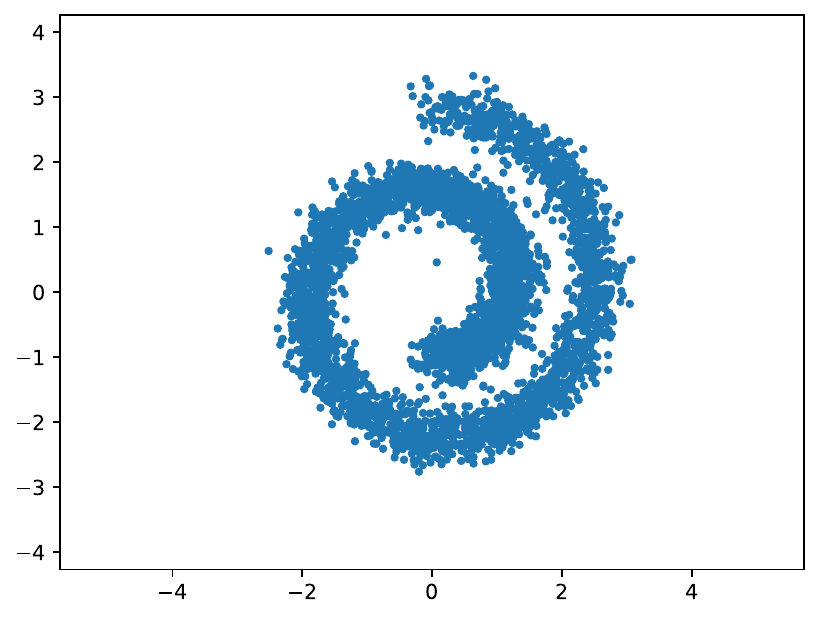}
    \end{minipage}
\hspace{-\samplehinterval}
    \begin{minipage}[t]{\samplempwid\textwidth}
    \centering
    \small{checkerboard}\\
    \includegraphics[width=\samplefigwid
    ]{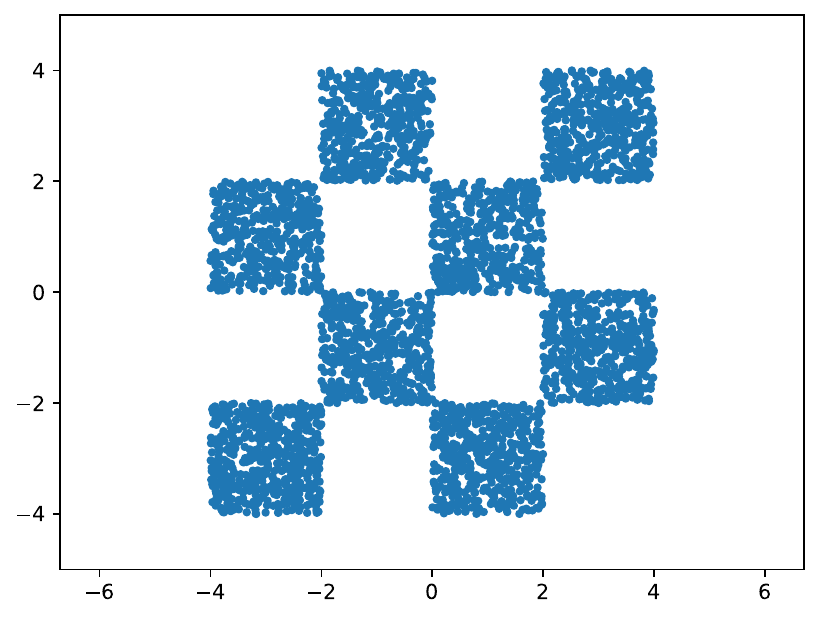}
    \end{minipage}
\hspace{-\samplehinterval}
\end{minipage}
\begin{minipage}{\textwidth}
    \begin{minipage}{\mpwid\textwidth}
    \centering
    \includegraphics[width=\textwidth,trim=0 30 0 30,clip
    ]{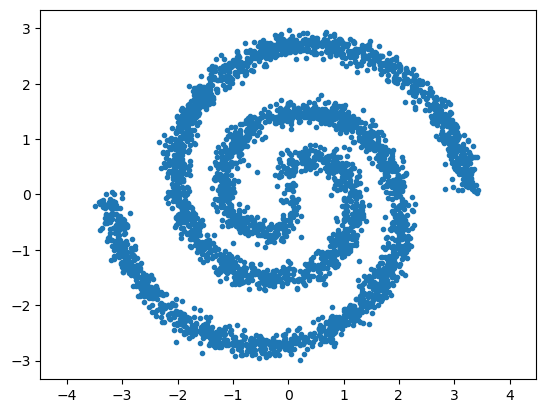}
    \end{minipage}
\hspace{-\hinterval}
    \begin{minipage}{\mpwid\textwidth}
    \centering
    \includegraphics[width=\textwidth,trim=0 30 0 30,clip
    ]{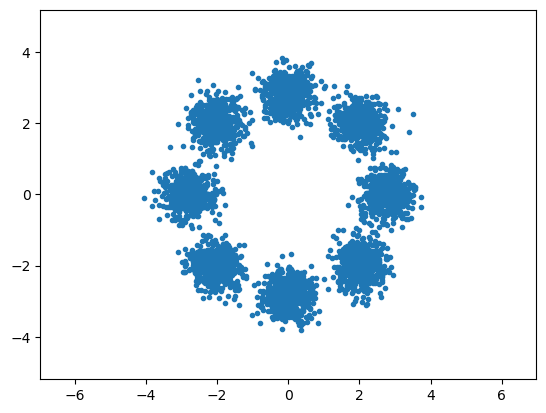}
    \end{minipage}
\hspace{-\hinterval}
    \begin{minipage}{\mpwid\textwidth}
    \centering
    \includegraphics[width=\textwidth,trim=0 30 0 30,clip
    ]{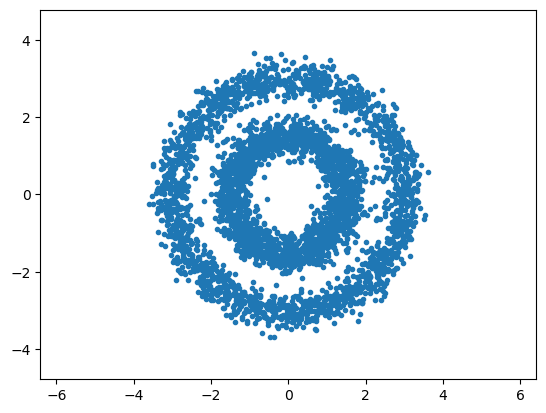}
    \end{minipage}
\hspace{-\hinterval}
    \begin{minipage}{\mpwid\textwidth}
    \centering
    \includegraphics[width=\textwidth,trim=0 30 0 30,clip
    ]{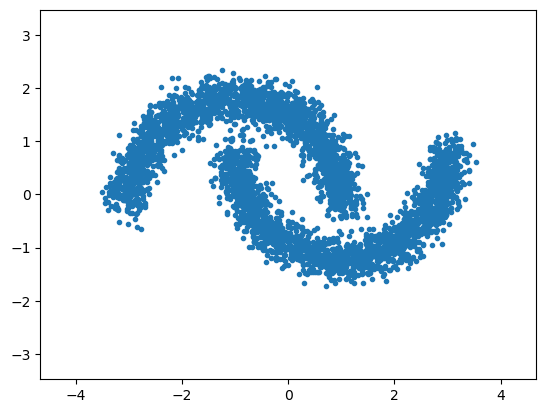}
    \end{minipage}
\hspace{-\hinterval}
    \begin{minipage}{\mpwid\textwidth}
    \centering
    \includegraphics[width=\textwidth,trim=0 30 0 30,clip
    ]{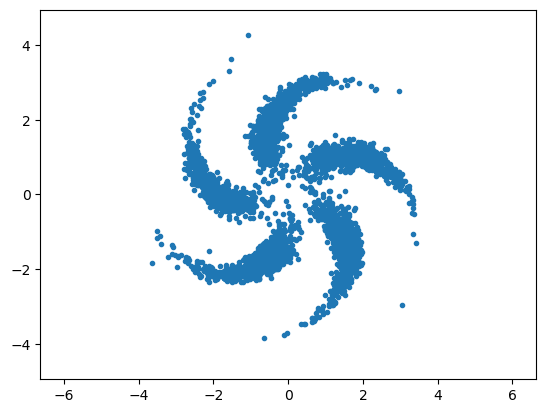}
    \end{minipage}
\hspace{-\hinterval}
    \begin{minipage}{\mpwid\textwidth}
    \centering
    \includegraphics[width=\textwidth,trim=0 30 0 30,clip
    ]{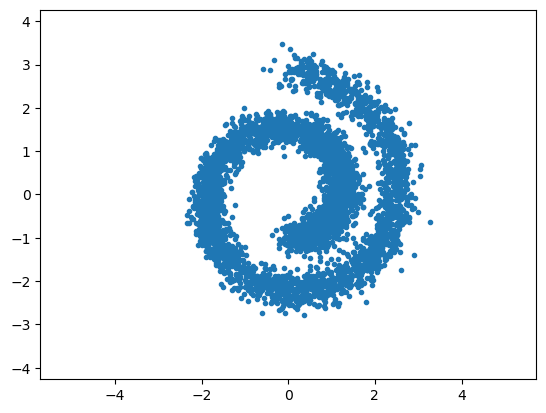}
    \end{minipage}
\hspace{-\hinterval}
    \begin{minipage}{\mpwid\textwidth}
    \centering
    \includegraphics[width=\textwidth,trim=0 30 0 30,clip
    ]{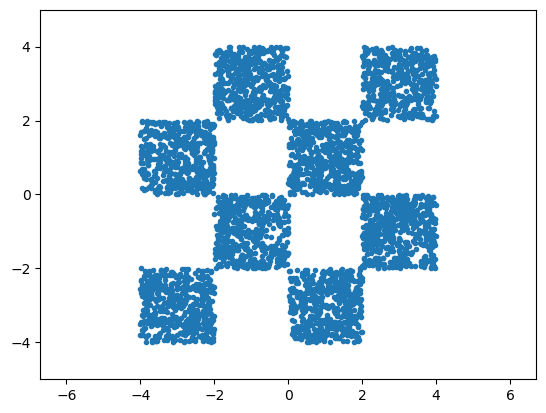}
    \end{minipage}
\hspace{-\hinterval}
\end{minipage}
\centering

\caption{
\emph{Top:} Visualization of the samples for synthetic problems from ground truth.
\emph{Bottom:} Visualization of the samples generated with the proposed GFlowNet-based algorithm.
}

\label{fig:gfn_synthetic_result}
\end{figure*}

\begin{table*}[t]
\setlength{\tabcolsep}{2.5mm}
\centering
\caption{
Experiment results with seven 2D synthetic problems.
We display the negative loglikelihood (NLL).
}
\label{tab:synthetic_nll}
\begin{tabular}{c|l|ccccccc}
\toprule
Metric & Method & 2spirals & 8gaussians & circles & moons & pinwheel & swissroll & checkerboard \\
\midrule
\multirow{4}{*}{NLL$\downarrow$} 
& PCD    &  $20.094$ & $19.991$ &$20.565$&$19.763$&$19.593$&$20.172$&$21.214$ \\
& ALOE   &  $20.295$ & $20.350$ & $20.565$ & $19.287$ & $19.821$ & $20.160$ & $54.653$ \\
& ALOE+  &  $20.062$ & $19.984$ & $20.570$ & $19.743$ & $19.576$ & $20.170$ & $21.142$\\
& EB-GFN & $\textbf{20.050}$ & ${19.982}$ & $\textbf{20.546}$ & ${19.732}$ & $\textbf{19.554}$ & $\textbf{20.146}$ & ${20.696}$ \\
& TBC-GFN & $\textbf{20.050}$ & $\textbf{19.965}$ & ${20.554}$ & $\textbf{19.719}$ & $\textbf{19.555}$ & $\textbf{20.144}$ & $\textbf{20.682}$ \\
\bottomrule
\end{tabular}

\end{table*}

We visualize the ground truth samples and GFlowNet generated samples in its $2$-D form in \Figref{fig:gfn_synthetic_result}. We also demonstrate the likelihood evaluation results in Table~\ref{tab:synthetic_nll}, which indicates that the proposed method achieves a fairly good level of distribution fitting. Regarding training details, we use the Adam optimizer with $1\times 10^{-4}$ and $1\times 10^{-5}$ for learning the forward and backward policy, respectively. The training keeps $100000$ steps. The evaluation of exponential MMD is calculated through average over $10$ sets of samples, each set contain $4000$ samples, which is the same as with~\citet{Zhang2022GenerativeFN}. All other setup follows \citet{Zhang2022GenerativeFN}, too.

\subsection{DDPM Demonstration}

\begin{figure}[t]
\centering
    \begin{minipage}{0.48\textwidth}
    \centering
    \includegraphics[width=\textwidth]{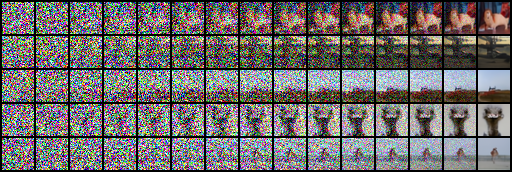}
    \end{minipage}
    \hspace{0.2cm}
    \begin{minipage}{0.22\textwidth}
    \centering
    \includegraphics[width=\textwidth]{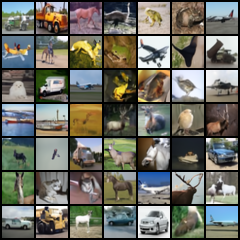}
    \end{minipage}
    \hspace{0.2cm}
    \begin{minipage}{0.22\textwidth}
    \centering
    \includegraphics[width=\textwidth]{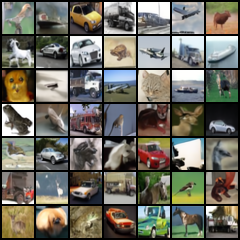}
    \end{minipage}
\caption{\textit{Left:} visualization of trajectory examples on CIFAR-$10$ image space. \textit{Right:}  MLE-GFN generated samples.}
\label{fig:ddpm_vis}
\end{figure}

We train on a single V100 GPU for $200$k steps, which takes less than three days.
We use the Adam optimizer with a learning rate of $2\times 10^{-4}$ for updating the forward policy and a learning rate of $2\times 10^{-5}$ for updating the backward policy. All NLL results are computed in the unit of bits per dim (BPD). The denoising process (forward policy) is parameterized with a UNet as done by \citet{Ho2020DenoisingDP}. The backward policy is parameterized in the way that its parameter $\vphi=\{\phi_i\}_i, \beta_i = \bar\beta_i\cdot\exp{(\phi_i)}$, where $\{\bar\beta_i\}_i$ are the original variance parameters from \citet{Ho2020DenoisingDP}.
We also visualize several trajectories of the GFlowNet in \Figref{fig:ddpm_vis} (left), where the intermediate state is updated with the forward policy from the left-hand side to the right-hand side of the figure. In \Figref{fig:ddpm_vis} (right), we visualize several generated image samples from MLE-GFN training on CIFAR-$10$ dataset.

\end{document}